\documentclass[10pt,twocolumn,letterpaper]{article}

\usepackage{iccv}
\newif\ificcvfinal
\iccvfinaltrue

\usepackage{graphicx}
\usepackage{amsmath,amssymb}
\usepackage{booktabs}
\usepackage{subcaption}
\usepackage{siunitx}
\usepackage{multirow}
\usepackage{tikz}
\usepackage{xfrac}
\usetikzlibrary{arrows.meta, positioning, shapes.misc, shapes.geometric, backgrounds, fit}
\usepackage[absolute,overlay]{textpos}

\definecolor{iccvblue}{rgb}{0.21,0.49,0.74}
\usepackage[pagebackref,breaklinks,colorlinks,allcolors=iccvblue]{hyperref}

\usepackage[capitalize]{cleveref}
\crefname{section}{Sec.}{Secs.}
\Crefname{section}{Section}{Sections}
\Crefname{table}{Table}{Tables}
\crefname{table}{Tab.}{Tabs.}

\def\paperID{9Tetfx3oHA}
\def\confName{ICCV}
\def\confYear{2025}

\title{Predictive Quality Assessment for Mobile Secure Graphics}
\author{
Cas Steigstra \\
Scantrust / University of Amsterdam \\
{\tt\small cas.steigstra@gmail.com}
\and
Sergey Milyaev \\
Scantrust \\
{\tt\small sergey.milyaev@gmail.com}
\and
Shaodi You \\
University of Amsterdam \\
{\tt\small s.you@uva.nl}
}

\begin{document}
\maketitle


\begin{textblock*}{\textwidth}[0.5, 0.5](0.5\paperwidth, 11.5cm)
\centering
\includegraphics[width=\linewidth]{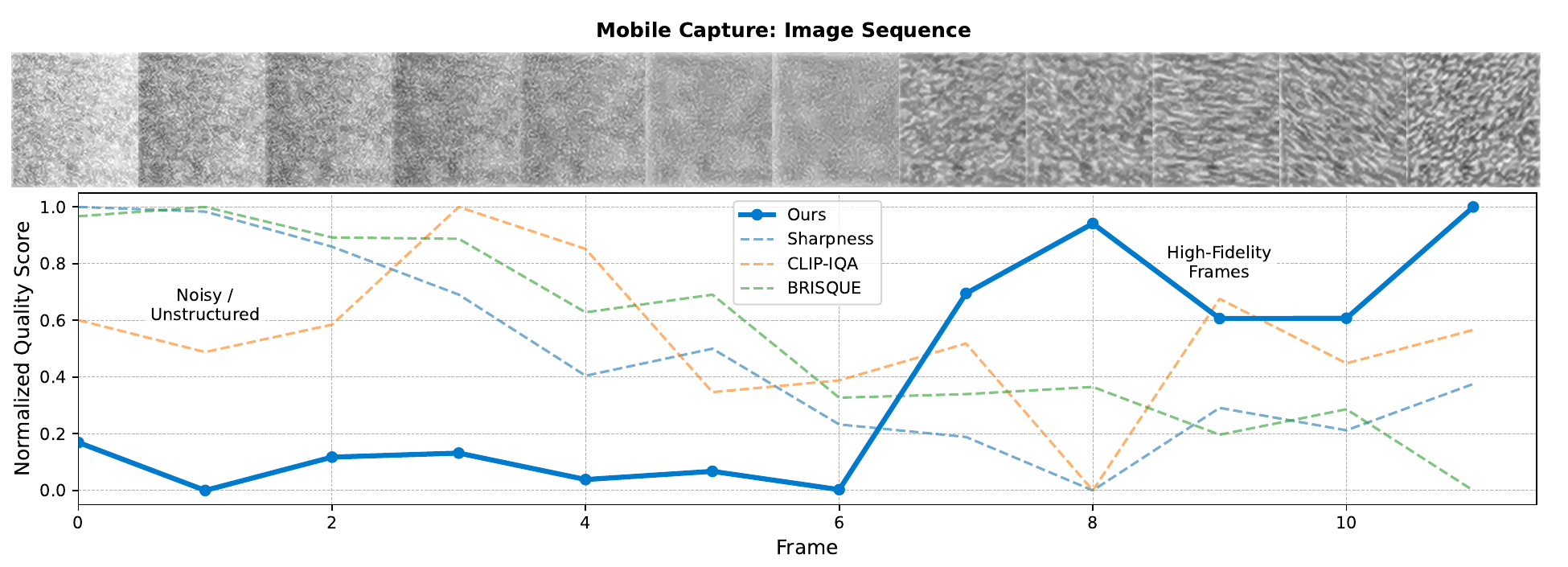}
\captionof{figure}{
    \textbf{Our Predictive Score Aligns with a Secure Graphic's Quality.}
    A mobile capture session progresses from noisy (left) to high-fidelity (right). Our predictive score (\textbf{Ours}, blue) correctly tracks this improvement. In contrast, standard IQA methods (dashed lines) prefer the initial noisy frames, giving a misleading signal about the frame's utility for verification. This shows the need for a task-specific approach over general-purpose IQA.
}
\label{fig:teaser}
\end{textblock*}
\vspace*{8.5cm}

\begin{abstract}
The reliability of secure graphic verification, a key anti-counterfeiting tool, is undermined by poor image acquisition on smartphones. Uncontrolled user captures of these high-entropy patterns cause high false rejection rates, creating a significant `reliability gap'. To bridge this gap, we depart from traditional perceptual IQA and introduce a framework that predictively estimates a frame's utility for the downstream verification task. We propose a lightweight model to predict a quality score for a video frame, determining its suitability for a resource-intensive oracle model. Our framework is validated using re-contextualized FNMR and ISRR metrics on a large-scale dataset of 32,000+ images from 105 smartphones. Furthermore, a novel cross-domain analysis on graphics from different industrial printing presses reveals a key finding: a lightweight probe on a frozen, ImageNet-pretrained network generalizes better to an unseen printing technology than a fully fine-tuned model. 

This provides a key insight for real-world generalization: 
\vspace*{8.5cm}

for domain shifts from physical manufacturing, a frozen general-purpose backbone can be more robust than full fine-tuning, which can overfit to source-domain artifacts.
\end{abstract}

\section{Introduction}
\label{sec:intro}

\begin{figure}[t]
\centering
\begin{tikzpicture}
\node[inner sep=0] (image) {\includegraphics[width=0.6\linewidth]{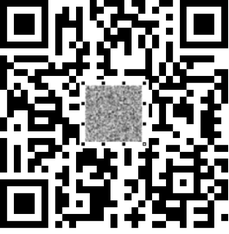}}; 

\begin{scope}[
remember picture,
overlay,
annot_text/.style={
font=\sffamily\small,
align=center,
fill=white, 
fill opacity=0.95, 
text opacity=1, 
draw=black,
rounded corners=2pt,
inner sep=2pt
},
arrow/.style={-{[width=6pt]Stealth}, thick}
]
\draw[red, thick, rounded corners=1pt]
([xshift=-0.62cm,yshift=-0.53cm]image.center)
rectangle
([xshift=0.58cm,yshift=0.68cm]image.center);

\node[annot_text, text=red] at ([yshift=-0.5cm]image.north) (SG_label) {Embedded Secure \\ Graphic (SG)};

\draw[arrow, red] (SG_label.south) -- ([yshift=0.7cm]image.center);

\draw[green, thick, rounded corners=1pt]
([xshift=-0.78cm,yshift=-0.54cm]image.center)
rectangle
([xshift=0.75cm,yshift=-0.65cm]image.center);

\node[annot_text, text=green] at ([yshift=-0.1cm,xshift=-0.1cm]image.west) (marker_label) {Synchronization \\ Markers};

\draw[arrow, green] (marker_label.east) -- ([yshift=-0.6cm,xshift=-0.78cm]image.center);

\node[annot_text, text=blue] at ([xshift=0.1cm]image.east) (qr_label) {Standard QR \\ Code Region \\ (Public Data)};

\draw[arrow, blue] (qr_label.south) -- ([xshift=-1.6cm, yshift=-1.5cm]image.east);

\end{scope}
\end{tikzpicture}
\caption{\textbf{Anatomy of the secure graphic used in this study.} The graphic combines a standard QR code (public data) with an embedded, high-entropy secure graphic (center). The secure graphic is an unclonable physical fingerprint, while the QR structure ensures mobile scanning compatibility.}
\label{fig:anatomy_of_graphic}
\vspace{-4mm}
\end{figure}

Counterfeit goods pose a formidable global challenge, inflicting trillions in economic damage and threatening public safety \cite{European2021Anti}. Physical-digital verification systems, which leverage the ubiquity of smartphones to verify physical products, have emerged as a promising solution. At the forefront of this are camera-verifiable \emph{security graphics}. As illustrated in \Cref{fig:anatomy_of_graphic}, our work focuses on a common implementation of this technology: a high-entropy secure graphic embedded within a standard QR code \cite{Picard2021Counterfeit, Tutt2024Authentication}. This hybrid design creates a secure, unclonable link to a digital identity that can be printed directly onto products.

Current research in security graphic verification has primarily focused on the security of the patterns themselves and the development of classifiers to distinguish genuine items from sophisticated fakes \cite{Chaban2021MachineLA, Khermaza2021Can, Belousov2024Machine}. These works, while crucial, almost universally operate under a strong, implicit assumption: that the captured image is of sufficient quality for reliable analysis. This task is highly sensitive to image quality, creating a ``minimum quality gap'' in real-world use. Our work addresses this foundational quality problem. Users capture images under uncontrolled conditions—poor lighting, motion blur, and defocus—leading to high false rejection rates \cite{Taran2023Mobile, Picard2008Copy}. More critically, severe image degradation can erode the statistical features of a genuine graphic to the point where it becomes indistinguishable from a counterfeit, turning a usability issue into a fundamental security vulnerability.
Resolving this gap is crucial for the widespread adoption of such visual verification systems. Our work addresses this challenge within the broader, emerging field of no-reference quality assessment for specialized, non-natural domains such as screen content and game graphics, where traditional models fail \cite{Zhu2025GameQA, Chen2021SCIDAQA}.

To bridge this minimum quality gap, this paper introduces a new paradigm: \emph{proactive, predictive quality assessment}. Instead of using low-level texture statistics for selection of an appropriate frame, which are computed independently of a verification model and may not be well correlated with it, our approach uses a light-weight model to predict the utility of a captured video frame for the downstream verification task. This model analyzes each frame from a live video stream in real-time and determines its likelihood of being successfully verified by a definitive, but potentially resource-intensive, oracle model \emph{before} it is ever submitted. The rationale for this two-stage approach is that the definitive oracle may be too computationally expensive for real-time processing (e.g., involving complex cryptographic checks) or may require a network round-trip to a secure server for verification. Our lightweight on-device filter thus makes the system practical by minimizing latency, server load, and power consumption. By proactively filtering out low-quality frames, our system dramatically reduces false rejections and can guide the user toward a successful scan.

The evaluation of the proposed module is inspired by the mature field of biometrics \cite{International2016ISO, grother2022frvt}. We design a lightweight, no-reference quality estimation model, $Q$, trained to directly regress the output of the definitive verification function. The performance of this system is measured through a principled trade-off analysis between the \textbf{False Non-Match Rate (FNMR)} and the \textbf{Incorrect Sample Rejection Rate (ISRR)} \cite{Grother2007Performance}. This methodology provides a holistic, system-level evaluation that quantifies the delicate balance between reliability and usability.

We validate our method on a large-scale, real-world dataset. Furthermore, to probe the limits of our model's robustness, we conduct a novel cross-domain analysis, testing it on graphics produced by an entirely unseen industrial printing process. This analysis reveals a key scientific finding: full network fine-tuning, while optimal in-domain, can overfit to the microscopic artifacts of a specific manufacturing process. We show that a more robust model is achieved by using a lightweight probe on a frozen, pre-trained feature backbone, providing a critical insight into building computer vision systems that generalize in the physical world.

The contributions of this paper are three-fold:
\begin{itemize}
\item We formulate the novel task of proactive, predictive quality assessment for the quality of security graphics, addressing the critical ``minimum quality gap'' in existing systems.
\item We introduce a principled, biometrics-inspired evaluation framework using FNMR/ISRR to holistically measure the system-level impact of a quality assessment model on both reliability and usability.
\item We provide the first empirical analysis of generalization across physical manufacturing domains, revealing that a lightweight probing approach can be more robust than full network fine-tuning.
\end{itemize}

\section{Related Work}
\label{sec:related_work}

\subsection{Secure Graphics}
The concept of using a high-entropy, copy-sensitive secure graphic for document security was pioneered by Picard \cite{Picard2004Digital}. The core principle is that any print-and-scan (P\&S) cycle introduces irreversible information loss \cite{Zhang2019Copy}, allowing an original to be distinguished from a copy \cite{Dirik2012Copy, Khermaza2021Can}. This secure graphic, often called Copy Detection Pattern (``CDPs''), leverage the stochastic nature of the printing process itself to create a unique physical fingerprint \cite{Tutt2022MathematicalMO, Chaban2024Comparative}. Early methods relied on classical image similarity metrics like correlation \cite{Picard2004Digital}, while later work explored more complex feature sets \cite{Dirik2012Copy, Zhang2019Copy, Taran2023Mobile} and sophisticated modeling of the print-scan channel \cite{Tutt2024Authentication}.

The move to smartphone-based verification has made this technology more accessible, often by embedding a secure graphic within standard 2D barcodes like QR codes \cite{Picard2021Counterfeit, Nguyen2019Watermarking, Tkachenko2016Two}. This accessibility increases scan quality variation from motion blur, defocus, or uneven lighting. Such degradations in principle increase information loss for the captured original samples. However, the current line of research in the CDP domain has primarily focused on bolstering security against advanced cloning and machine learning-based attacks \cite{Chaban2021MachineLA, Belousov2024Machine}. This research often assumes high-quality acquisition, using lab scanners or manually curating mobile captures \cite{Chaban2024Comparative, Tutt2024Authentication}, which sidesteps the real-world variance problem we address. A production secure graphic authentication system, like the one presented in \cite{Picard2021Counterfeit}, can't store the reference data required for verification on mobile devices due to security concerns. That makes frame-by-frame on-device verification impossible and requires frame selection based on secure graphic capture quality. To minimize the usage of the mobile network, only selected frames are sent to the authentication server for verification. Our model provides quality estimates to be used for the best frame selection to maximize the chance of successful authentication on the server.

\begin{figure}[t]
\centering
\includegraphics[width=\linewidth]{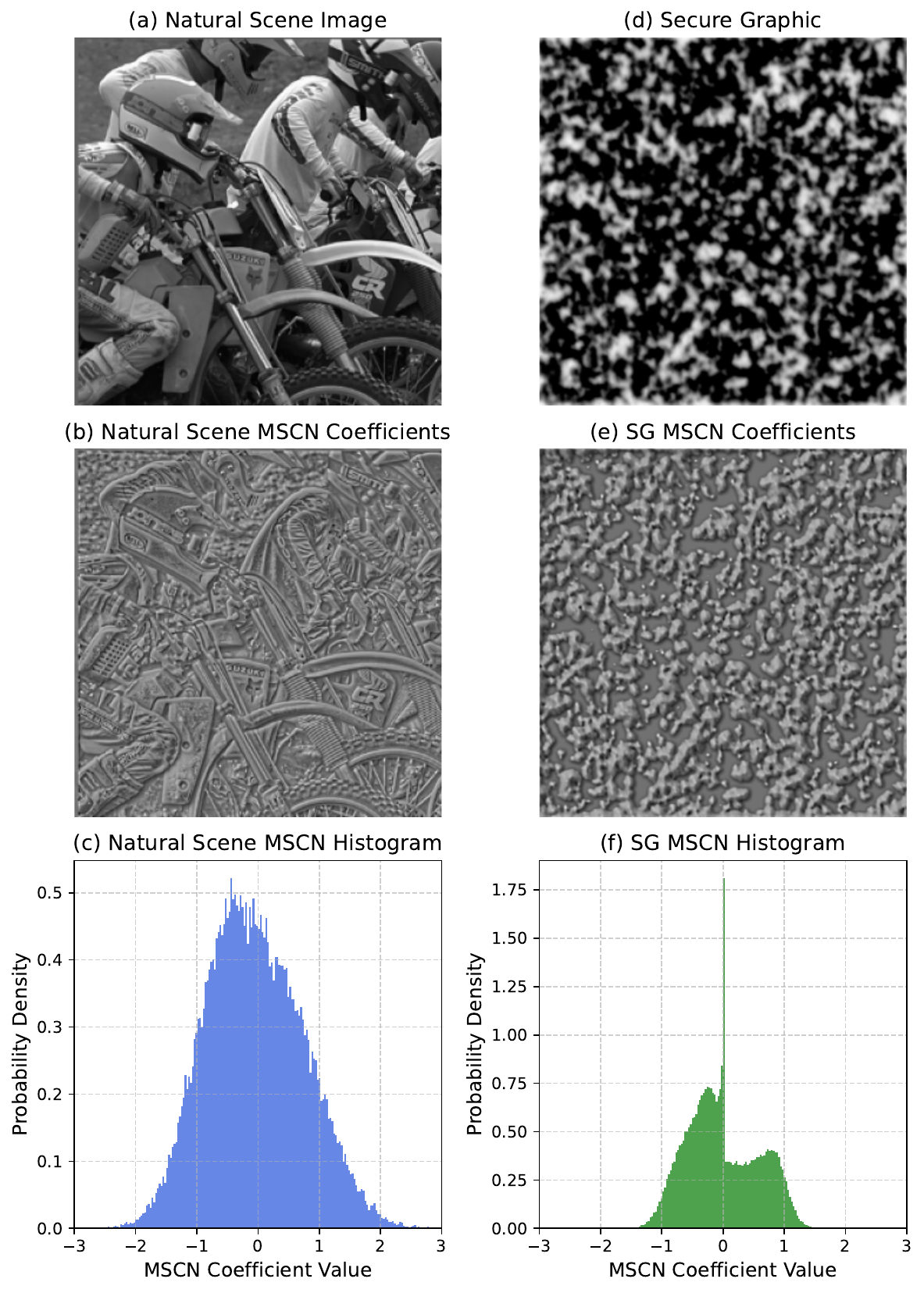}
\caption{
\textbf{Why Natural Scene Statistics (NSS) Do Not Apply to Security Graphics.}
Contrast between a natural scene (left \cite{sheikh2006statistical}) and our high-entropy graphic (right \cite{yadav2019estimation}). The natural scene's MSCN histogram is unimodal and near-Gaussian, conforming to the NSS assumption of standard NR-IQA models. Our graphic's histogram is bimodal, fundamentally violating this assumption and showing why standard IQA is unsuited for our task.
}
\label{fig:nss_mismatch}
\vspace{-4mm}
\end{figure}

\subsection{Image Quality Assessment (IQA)}
Our work is situated within the field of No-Reference Image Quality Assessment (NR-IQA), as the on-device model must assess quality without access to a pristine reference \cite{Yang2019Survey, Wang2021survey}. NR-IQA research can be broadly divided into two paradigms.

\paragraph{Perceptual IQA.} The dominant paradigm focuses on predicting human-perceived aesthetic quality. Many classical methods, such as BRISQUE \cite{Mittal2012No} and NIQE \cite{Mittal2013Niqe}, are based on the Natural Scene Statistics (NSS) hypothesis, which posits that natural images have predictable statistical properties that are distorted by noise or compression \cite{Sheikh2005information}. Deep learning has also been extensively applied to learn features that correlate with human opinion scores \cite{Kang2014Convolutional, Bosse2016Deep}. Application of foundation vision models for IQA is an active area of research \cite{Zhang2024QualityAI}, but they can't be applied to real-time frame processing on mobile devices. These perceptual approaches are fundamentally mismatched for our task. As illustrated in \Cref{fig:nss_mismatch}, a secure graphic contains high-entropy, non-NSS patterns whose statistics violate the core assumptions of these models. Our goal is to predict machine-verifiability, not human aesthetic preference. The potential of finetuning a pre-trained model for perceptual IQA was demonstrated in \cite{Talebi2018NIMA}. In our work, we explore various strategies for model finetuning, taking into account the bigger gap between the source and target domains for the selected pre-trained model.

\paragraph{Predictive IQA.} A more relevant paradigm, primarily developed in biometrics, defines quality by a sample's utility for a downstream recognition task \cite{schlett2022face}. For instance, the quality model for a face image is evaluated by how it affects the probability of a false match or non-match \cite{Grother2007Performance, International2016ISO, grother2022frvt}. This task-based definition of quality is what we adopt. We are the first to apply this predictive paradigm to physical secure graphic verification, developing a quality model whose objective function is directly tied to the success of the core security task. A sample's utility for a document recognition system was also explored in \cite{alaei2023document} by connecting an image document quality metric to OCR accuracy. In our scenario, each image sample has a single prediction, so the estimated accuracy for it is always a binary value and can't serve as a proxy for a utility measure. The frame quality assessment for object detection was explored in \cite{Beniwal2023ImageQA}, where the proposed IQA assumes only degradations from image compression and is calculated from intermediate outputs of the selected CNN architecture for object detection. In contrast, our proposed model only requires the confidence output of a black-box classifier, needing no intermediate representations \cite{Beniwal2023ImageQA} or specific task formulations \cite{schlett2022face, alaei2023document}.

\section{Problem Formulation}
\label{sec:problem_formulation}

We formally define the verification pipeline, the ``minimum quality gap'' challenge, and our proposed solution.
\subsection{Standard Verification Pipeline}
Let $I_{SG} \in \{0,1\}^{H \times W}$ be the pristine, high-entropy binary digital reference of a security graphic. The \textbf{genuine} physical instance results in a captured image $I_S \in \mathbb{R}^{H' \times W'}$ (after geometric alignment). The downstream verification process compares the scan $I_S$ to its digital reference $I_{SG}$, yielding a continuous \textbf{quality score}, $s$:
\begin{equation}
    s = M(I_S, I_{SG}),
    \label{eq:verification_function}
\end{equation}
where a higher score indicates greater fidelity. A genuine scan is \textbf{high quality} if its score meets or exceeds a predefined threshold, $\tau$. This partitions the set of all genuine scans, $\mathcal{I}_{\text{genuine}}$, into two disjoint subsets:
\begin{align}
    \mathcal{I}_{\text{high-quality}} &= \{ I_S \in \mathcal{I}_{\text{genuine}} \mid M(I_S, I_{SG}) \geq \tau \}, \\
    \mathcal{I}_{\text{low-quality}} &= \{ I_S \in \mathcal{I}_{\text{genuine}} \mid M(I_S, I_{SG}) < \tau \}.
\end{align}
The large set of genuine but poor-quality scans, $\mathcal{I}_{\text{low-quality}}$, is the core problem, causing high false rejection rates.

\subsection{Proactive Quality Assessment as a Surrogate Model}
To prevent the processing of scans from $\mathcal{I}_{\text{low-quality}}$, we propose learning a lightweight, no-reference \emph{surrogate model}, $Q$. This model's goal is to predict a score $q = Q(I_S)$ that accurately estimates the oracle score $s$:
\begin{equation}
Q(I_S) \approx s = M(I_S, I_{SG}).
\label{eq:surrogate_model}
\end{equation}
By applying a quality threshold, $\sigma$, it can act as an intelligent gatekeeper, only selecting frames that are highly likely to be of high quality (i.e., where $Q(I_S) \geq \sigma$). This proactive filtering bridges the minimum quality gap.

\section{Proposed Methodology}
\label{sec:methodology}
This section details our methodology, including the system-level evaluation framework, the large-scale dataset, and the model architectures and training protocol.

\subsection{System-Level Evaluation Framework}
A quality model's efficacy lies in its impact on the end-to-end verification system. We evaluate our models using a framework adapted from biometrics \cite{Grother2007Performance, grother2022frvt, Schlett2023Considerations}.

We define a `match' as a scan that is of high quality and suitable for verification (i.e., $M(I_S, I_{SG}) \geq \tau$), and a `non-match' as a low-quality scan that is not ($M(I_S, I_{SG}) < \tau$). The performance of our quality model $Q$ is measured by the trade-off between two key error rates, calculated by sweeping a quality threshold $\sigma$ over the predicted scores $q = Q(I_S)$:

Note the two thresholds: $\tau$ is a fixed oracle security threshold used once to label the dataset into ground-truth high/low-quality sets. In contrast, $\sigma$ is the variable operating point for our surrogate model $Q$. We evaluate $Q$ by sweeping $\sigma$ across its full range to trace the error trade-offs:

\begin{itemize}
    \item \textbf{False Non-Match Rate (FNMR):} The rate at which $Q$ incorrectly accepts a low-quality sample ('non-match'), measuring gatekeeper failure.
    \begin{equation}
        \label{eq:fnmr}
        \text{FNMR}(\sigma) = \frac{|\{ I_S \in \mathcal{I}_{\text{low-quality}} \mid Q(I_S) \geq \sigma \}|}{|\{ I_S \mid Q(I_S) \geq \sigma \}|}.
    \end{equation}

    \item \textbf{Incorrect Sample Rejection Rate (ISRR):} The rate at which $Q$ incorrectly rejects a high-quality sample ('match'), measuring user friction.
    \begin{equation}
        \label{eq:isrr}
        \text{ISRR}(\sigma) = \frac{|\{ I_S \in \mathcal{I}_{\text{high-quality}} \mid Q(I_S) < \sigma \}|}{|\mathcal{I}_{\text{high-quality}}|}.
    \end{equation}
\end{itemize}

We visualize this trade-off using \textbf{Error versus Discard Characteristic (EDC) curves}, which plot FNMR or ISRR against the fraction of total samples discarded. To provide a single scalar metric for model comparison, we compute the partial Area Under the Curve (pAUC) and compare it against a theoretical \textbf{Ideal Observer} ($Q_{\text{Ideal}} = M$), which represents the best possible performance.

\subsection{Dataset for Validation}
\begin{figure}[h]
\centering
\includegraphics[width=\linewidth]{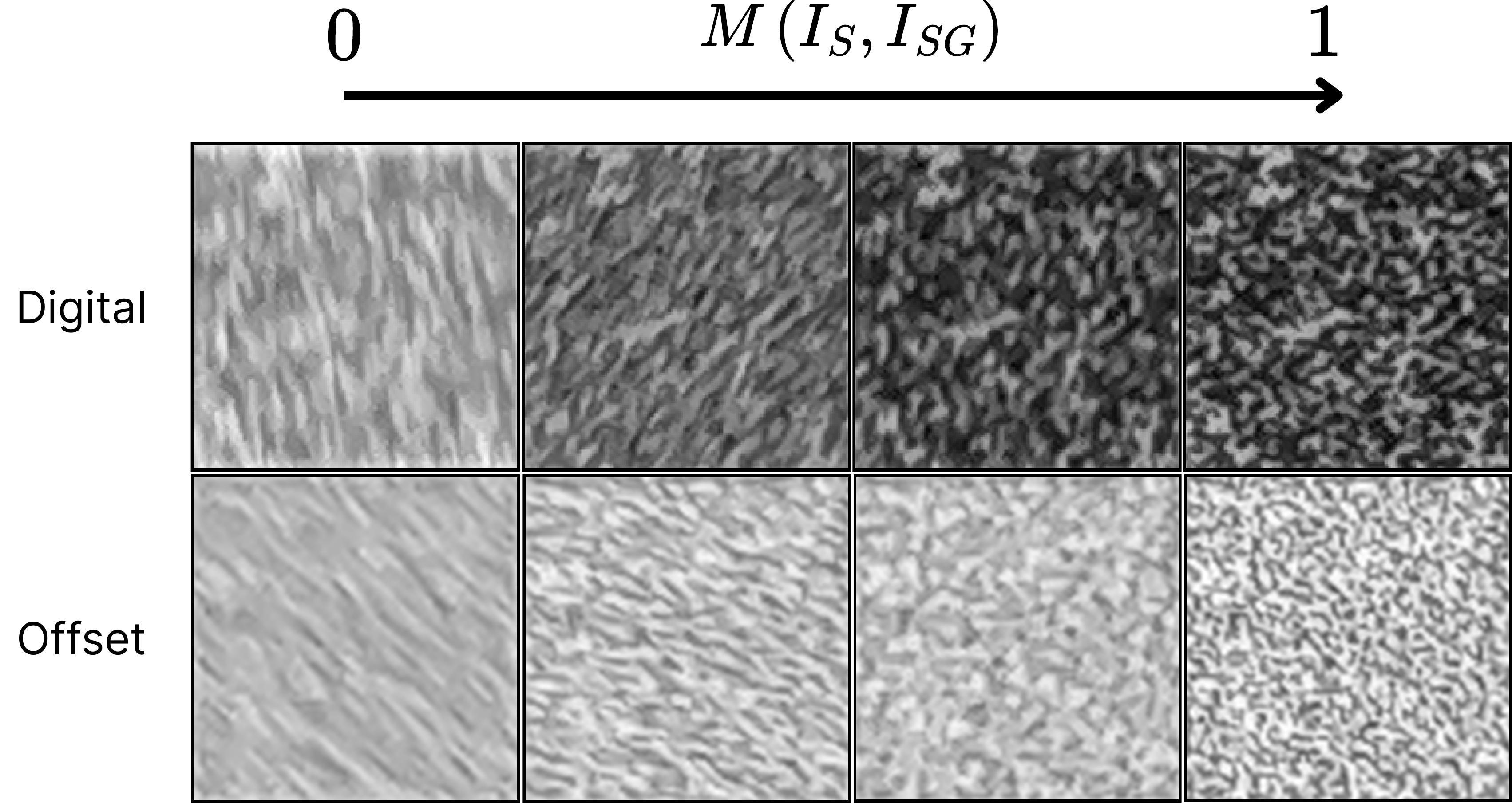}
\caption{
    \textbf{Visual Definition of the Quality Score Across Two Print Domains.}
    Security graphic samples sorted by their quality score $s$ (low to high fidelity). Low scores correspond to degraded scans (blur, low contrast), while high scores represent sharp patterns. The figure also shows textural differences between the two manufacturing processes (\textbf{Digital} vs. \textbf{Offset}), illustrating the physical domain shift challenge.
}
\label{fig:score_and_domain_examples}
\end{figure}

We use a large-scale industrial dataset curated for two key challenges: diverse \textbf{acquisition quality} and varied \textbf{physical manufacturing processes}. \Cref{fig:score_and_domain_examples} visually defines these dimensions, showing how score $s$ maps to degradation and how patterns differ between print domains (\textbf{Digital} vs. \textbf{Offset}).

\vspace{-.2cm}
\paragraph{Primary (In-Domain) Dataset.} This dataset was created using graphics printed with an industrial \textbf{HP Indigo digital press}. It contains \textbf{32,391} labeled image frames from 1,672 scan sessions, captured across \textbf{105 unique smartphone models} in uncontrolled, real-world conditions. This diversity is crucial for training a robust model.

\vspace{-.2cm}
\paragraph{Cross-Domain Dataset.} To test generalization, we use a second dataset of 5,104 images from \textbf{94 unique phones}, where graphics were printed using a traditional \textbf{offset printing press}. This introduces a significant domain shift rooted in a different physical manufacturing process and is used exclusively for testing generalization.

To ensure an unbiased evaluation, the dataset was rigorously partitioned at the level of the unique \textbf{physical graphic}, guaranteeing that all frames from a single graphic belong to only one split (train/val/test).

\subsection{Model Architectures and Training}
We evaluated a wide spectrum of architectures for the surrogate model $Q$. We detail our key baseline heuristics and our final proposed end-to-end model below. An overview of all evaluated paradigms is in \Cref{tab:model_overview_final}; full details are in the supplement.

\subsubsection{Key Evaluated Models}

\paragraph{Baseline Heuristics.} To establish a performance floor, we implemented two feature-based heuristics. The \textbf{Sharpness} metric quantifies quality by analyzing the magnitude of image gradients based on established principles of edge profile analysis \cite{Pech2000}. The \textbf{Blur} metric, instead of using a generic method like variance of the Laplacian \cite{Ye2013Real}, uses a specialized approach that leverages the QR code's structure to measure the effective edge width.

\vspace{-.2cm}
\paragraph{Hypothesis on NSS-Based Methods.} A core hypothesis of this work is that standard IQA methods based on Natural Scene Statistics (NSS), such as \texttt{BRISQUE} \cite{Mittal2012No}, are theoretically unsuited for this task. These methods assume that high-quality images share the statistical properties of natural scenes. Our secure graphics, being high-entropy, non-natural patterns, fundamentally violate this assumption, as illustrated in \Cref{fig:nss_mismatch}. We therefore expect these general-purpose models to fail, necessitating the task-specific approaches evaluated next.

\vspace{-.2cm}
\paragraph{Proposed End-to-End Model.} Our main proposal, \texttt{MobileNet\textsuperscript{IN}(SG|M)}, leverages the highly efficient \texttt{MobileNetV2} architecture \cite{Sandler2018MobileNetV2, Talebi2018NIMA}, pre-trained on ImageNet. We replace the original top classification layer with a new regression head composed of two fully-connected layers, culminating in a single neuron with a linear activation. The entire network—both the pre-trained convolutional base and the new regression head—is then fine-tuned on our CDP dataset.

\vspace{-.2cm}
\paragraph{Training Protocol.} All supervised models were trained to minimize the \textbf{Mean Squared Error (MSE)} between the predicted score $Q(I_S)$ and the ground-truth server score $M(I_S, I_{SG})$. Deep learning models were trained using the Adam optimizer with a learning rate of $1 \times 10^{-4}$. We employed early stopping based on validation set performance to select the best model checkpoint.

\begin{table}[h!]
\centering
\caption{Complete Overview of Evaluated Model Paradigms. Full details are provided in the supplementary material.}
\label{tab:model_overview_final}
\resizebox{\columnwidth}{!}{%
\begin{tabular}{@{}lll@{}}
\toprule
\textbf{Category} & \textbf{Model ID} & \textbf{Core Principle \& Citation} \\
\midrule
\multirow{2}{*}{\textit{Baselines}} & \texttt{Random} & No-information random score. \\
& \texttt{Sharpness} / \texttt{Blur} & Heuristics based on edge profile analysis \cite{Pech2000, Ye2013Real}. \\
\addlinespace
\multirow{2}{*}{\textit{General-Purpose IQA}} & \texttt{BRISQUE\textsuperscript{NSS}} & Measures deviation from Natural Scene Statistics \cite{Mittal2012No}. \\
& \texttt{CLIP-IQA} & Zero-shot query of a vision-language model \cite{Radford2021Learning}. \\
\addlinespace
\multirow{2}{*}{\textit{Unsupervised (Task-Adapted)}} & \texttt{NIQE (SG)} & Distance to a model of high-quality CDPs \cite{Mittal2013Making}. \\
& \texttt{NIQE-LBP (SG)} & NIQE using LBP texture features instead of NSS features. \\
\addlinespace
\multirow{2}{*}{\textit{Supervised (Handcrafted)}} & \texttt{BRISQUE (SG|M)} & Retrains BRISQUE features to predict the score $M$. \\
& \texttt{LBP (SG|M)} & Trains an SVR on LBP features to predict the score $M$. \\
\addlinespace
\multirow{3}{*}{\textit{Supervised (End-to-End)}} & \texttt{CNN-3x32 (SG|M)} & Learns features from scratch with a shallow CNN \cite{Kang2014Convolutional}. \\
& \texttt{MobileNet (SG|M)} & Trains \texttt{MobileNetV2} architecture from scratch \cite{Sandler2018MobileNetV2}. \\
& \texttt{MobileNet\textsuperscript{IN}(SG|M)} & Fine-tunes a pre-trained \texttt{MobileNetV2} (\textbf{Our Proposal}). \\
\bottomrule
\end{tabular}%
}
\end{table}

\begin{figure}[t]
\centering
\includegraphics[width=\linewidth]{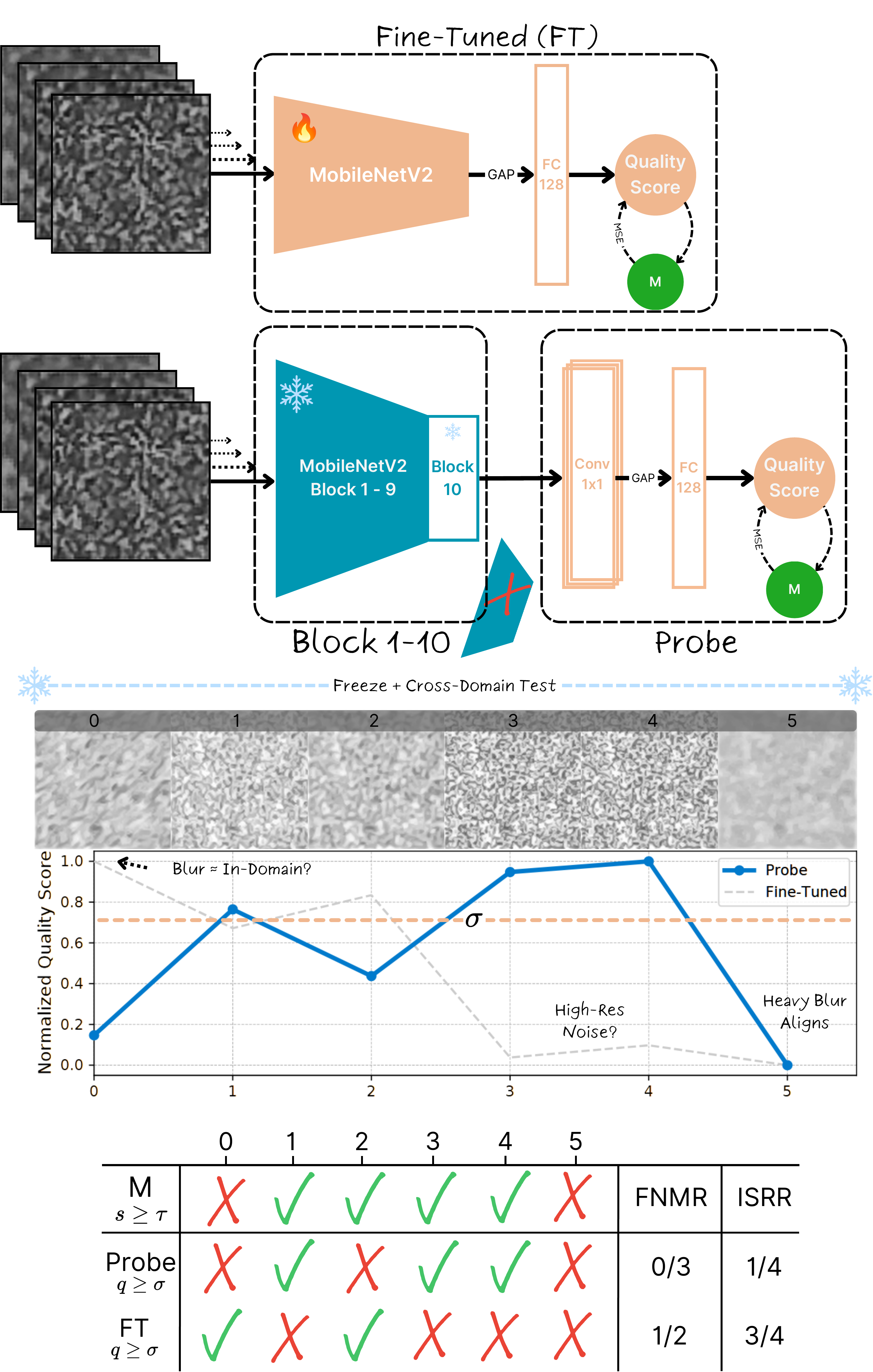} 
\caption{
    \textbf{Probing a Frozen Backbone Generalizes Better Than Full Fine-Tuning.}
    Comparing full network \textbf{fine-tuning (FT)} (top) with a lightweight \textbf{probe} on a frozen backbone (middle). The plot and table below show an example where the FT model incorrectly accepts a defective frame (\#0) that mimics source-domain artifacts. The probe correctly rejects it, demonstrating better robustness by avoiding the FT model's overfitting. This illustrates our key finding on generalization.
}
\label{fig:probing_methodology}
\vspace{-2mm}
\end{figure}

\begin{figure*}[t!]
\centering
\includegraphics[width=0.95\textwidth]{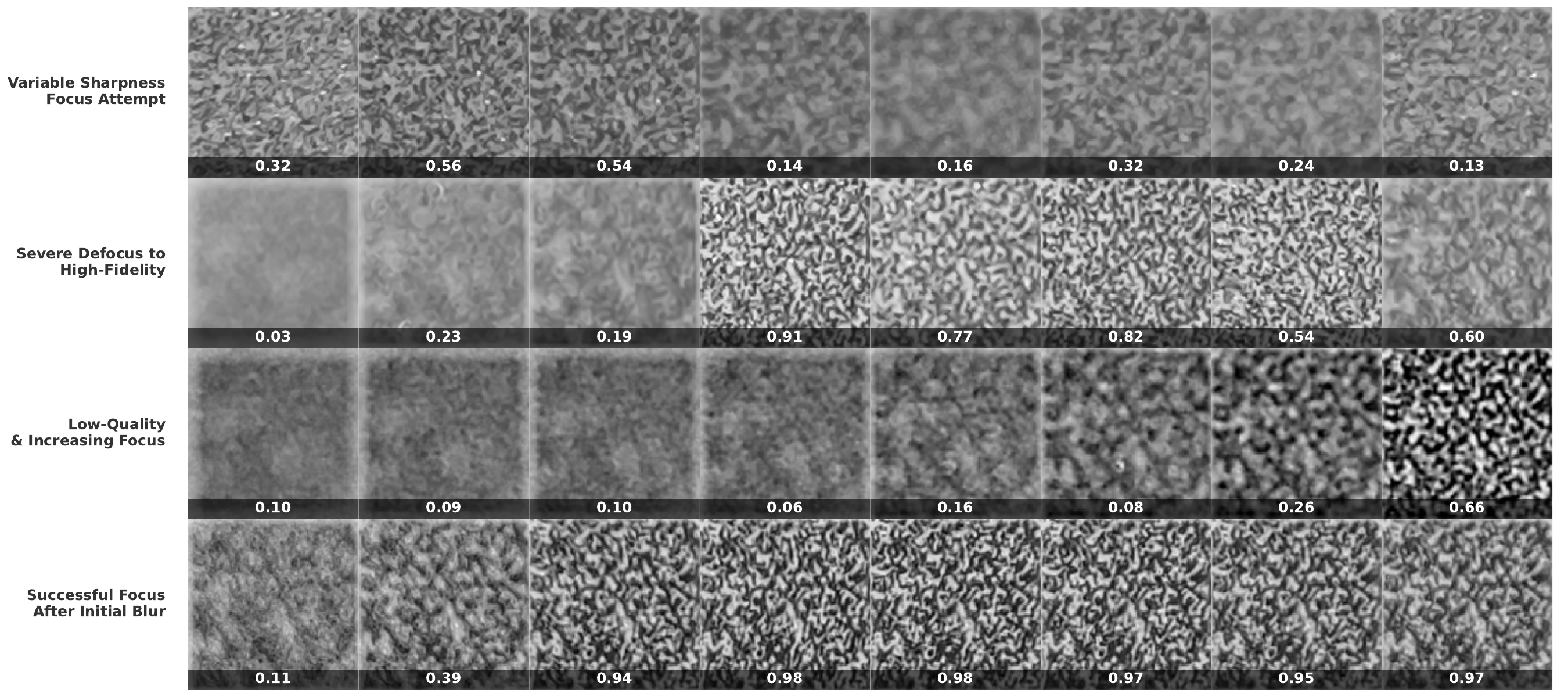}
\vspace{-2mm}
\caption{
\textbf{Visual Demonstration of Predicted Quality During Scanning Events.} Each row depicts a common scanning scenario, with frames shown in temporal order. Our model's predicted score (bottom) accurately tracks the visual quality, assigning low values to blurry or out-of-focus frames and high values to sharp, verifiable ones, effectively narrating the focus acquisition process.
}
\label{fig:ex1_visual}
\vspace{-2mm}
\end{figure*}

\subsubsection{Probing Analysis for Generalization}
A key challenge for real-world deployment is generalizing to unseen physical variations, such as different printing technologies. To investigate this, we employ a probing analysis to dissect the internal representations of our \texttt{MobileNet\textsuperscript{IN}(SG|M)} model and test a critical hypothesis about generalization.

As illustrated in \Cref{fig:probing_methodology}, this technique involves using the powerful \texttt{MobileNetV2} as a \textbf{frozen}, non-trainable feature extractor. We then attach a small, lightweight regression model, or \textbf{probe}, to its intermediate layers. By training \textit{only the probe's weights}, we can assess the predictive quality of features at different levels of abstraction. We hypothesize that while full fine-tuning is optimal in-domain, it risks overfitting to microscopic physical artifacts of the source domain. A probe, by leveraging more general features from the frozen backbone, may prove more robust. This analysis reveals the trade-off between specialization and generalization, a central theme of our results.


\section{Experiments}
\label{sec:experiments}

We present experiments structured to first visually and quantitatively validate our quality assessment framework, and then to investigate model robustness to a physical domain shift.

\subsection{Visual Verification of Quality Prediction}

We first visually validate the ability of our best model, \texttt{MobileNet\textsuperscript{IN}(SG|M)}, to track frame utility during a scan. \Cref{fig:ex1_visual} illustrates this by presenting four common scanning scenarios, with each frame annotated by our model's predicted quality score.

The model's scores effectively narrate the user's focus acquisition process:
\begin{itemize}
    \item During an initial \textbf{focus attempt} (Row 1), the scores correctly reflect the variable and often mediocre sharpness of the frames.
    \item The model dynamically tracks \textbf{increasing focus} (Rows 2 \& 3), showing a distinct and rapid transition. It assigns extremely low scores to severely defocused frames before jumping to high scores once fidelity is achieved.
    \item For a \textbf{successful scan} (Row 4), the model confirms the acquisition of a high-quality capture by assigning sustained high scores after overcoming an initial blur.
\end{itemize}

This demonstrates that our model operates beyond simple static quality estimation. It functions as an intelligent, real-time gatekeeper, ideal for identifying the optimal frame for verification in a live video stream.

\subsection{Quantitative Performance}
\paragraph{Purpose.} To validate our framework, systematically evaluate different modeling paradigms, and show the unsuitability of general-purpose IQA methods for this task.
\vspace{-.2cm}
\paragraph{Setup.} We evaluate representative models from each paradigm, detailed in \Cref{tab:model_overview_final}, on the held-out test set from our primary (Digital Press) dataset. Performance is measured using the $\Delta$pAUC of the FNMR and ISRR curves, representing the ``excess error'' over the Ideal Observer. Full, unabridged results for all models are provided in the supplementary material.

\begin{figure}[h]
\centering
\includegraphics[width=\linewidth]{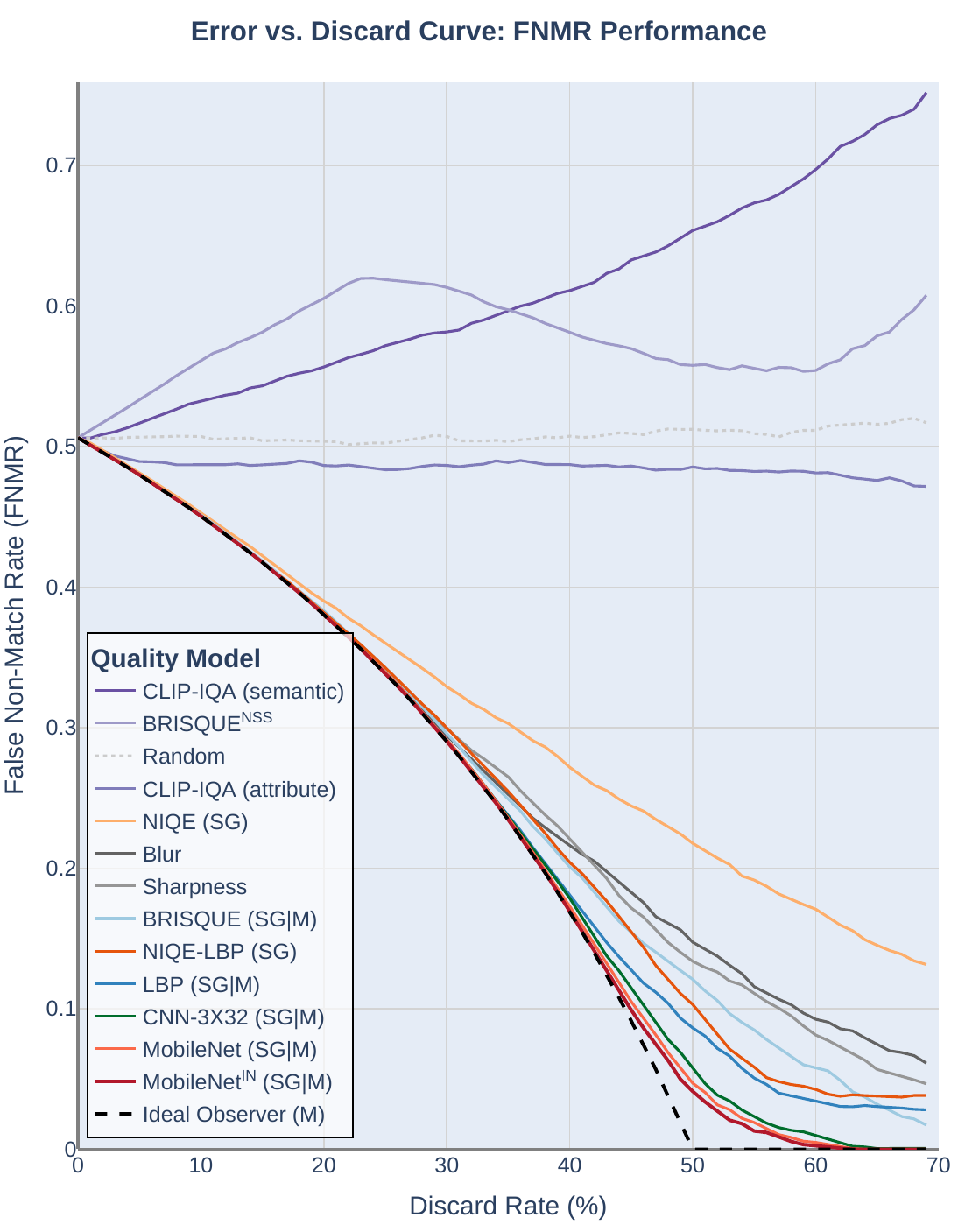}
\vspace{-2mm}
\caption{
\textbf{End-to-End Fine-Tuning Outperforms All Other Paradigms.}
Error versus Discard Characteristic (EDC) curves for the False Non-Match Rate (FNMR) on the in-domain test set. A lower curve indicates better performance, with the `Ideal Observer' representing the empirical optimum. The plot visually confirms the performance hierarchy of the different modeling paradigms; \textbf{the legend is sorted by FNMR $\Delta$pAUC to reflect this ranking.} Our proposed model (fine-tuned MobileNetV2) significantly outperforms all other approaches and closely tracks the ideal performance bound.
}
\label{fig:in_domain_edc}
\vspace{-2mm}
\end{figure}

\paragraph{Results.} \Cref{tab:exp1_pauc_final} shows a clear performance hierarchy. General-purpose models like \texttt{CLIP-IQA} fail, performing worse than random. Supervised models significantly outperform all others, and our proposed \texttt{MobileNetV2} is best, with its EDC curve (\Cref{fig:in_domain_edc}) closely tracking the Ideal Observer. This confirms an end-to-end approach learns an accurate quality surrogate.

As shown in recent work \cite{Li2023Rethinking}, the \texttt{MobileNetV2} model with 3.5M parameters and 0.3 GMACs has a latency of 1 ms on an iPhone 12 device with NPU inference, and 25 ms on a Pixel 6 phone with CPU inference. It proves that our proposed model is suitable for real-time on-device frame processing.

\begin{table}[htbp!]
\centering
\caption{In-Domain Performance: Partial Area Under the Curve ($\Delta$pAUC) for FNMR and ISRR over the 0-70\% discard rate range. Lower values are better. The best-performing result is in \textbf{bold}, with the second-best in \textit{italics}.}
\label{tab:exp1_pauc_final}
\footnotesize 
\setlength{\tabcolsep}{4pt} 
\begin{tabular}{@{}lcc@{}}
\toprule
\textbf{Method} & \textbf{FNMR $\Delta$pAUC~$\downarrow$} & \textbf{ISRR $\Delta$pAUC~$\downarrow$} \\
\midrule
\textit{Baselines} & & \\
\texttt{Random} & 0.2826 & 0.2967 \\
\texttt{Sharpness} & 0.0412 & 0.0398 \\
\texttt{Blur} & 0.0441 & 0.0414 \\
\addlinespace
\textit{General-Purpose IQA} & & \\
\texttt{BRISQUE\textsuperscript{NSS}} & 0.3504 & 0.3862 \\
\texttt{CLIP-IQA (Semantic)} & 0.3810 & 0.4003 \\
\texttt{CLIP-IQA (Attribute)} & 0.2601 & 0.2701 \\
\addlinespace
\textit{Unsupervised (Task-Adapted)} & & \\
\texttt{NIQE (SG)} & 0.0841 & 0.0828 \\
\texttt{NIQE-LBP (SG)} & 0.0273 & 0.0270 \\
\addlinespace
\textit{Supervised (Handcrafted)} & & \\
\texttt{BRISQUE (SG|M)} & 0.0301 & 0.0295 \\
\texttt{LBP (SG|M)} & 0.0185 & 0.0173 \\
\addlinespace
\textit{Supervised (End-to-End)} & & \\
\texttt{CNN-3x32 (SG|M)} & 0.0086 & 0.0086 \\
\texttt{MobileNet (SG|M)} & \textit{0.0063} & \textit{0.0064} \\
\texttt{MobileNet\textsuperscript{IN}(SG|M)} & \textbf{0.0042} & \textbf{0.0042} \\
\bottomrule
\end{tabular}
\end{table}

\subsection{Generalization to Physical Domain Shift}
\begin{figure}[h]
\centering
\includegraphics[width=\linewidth]{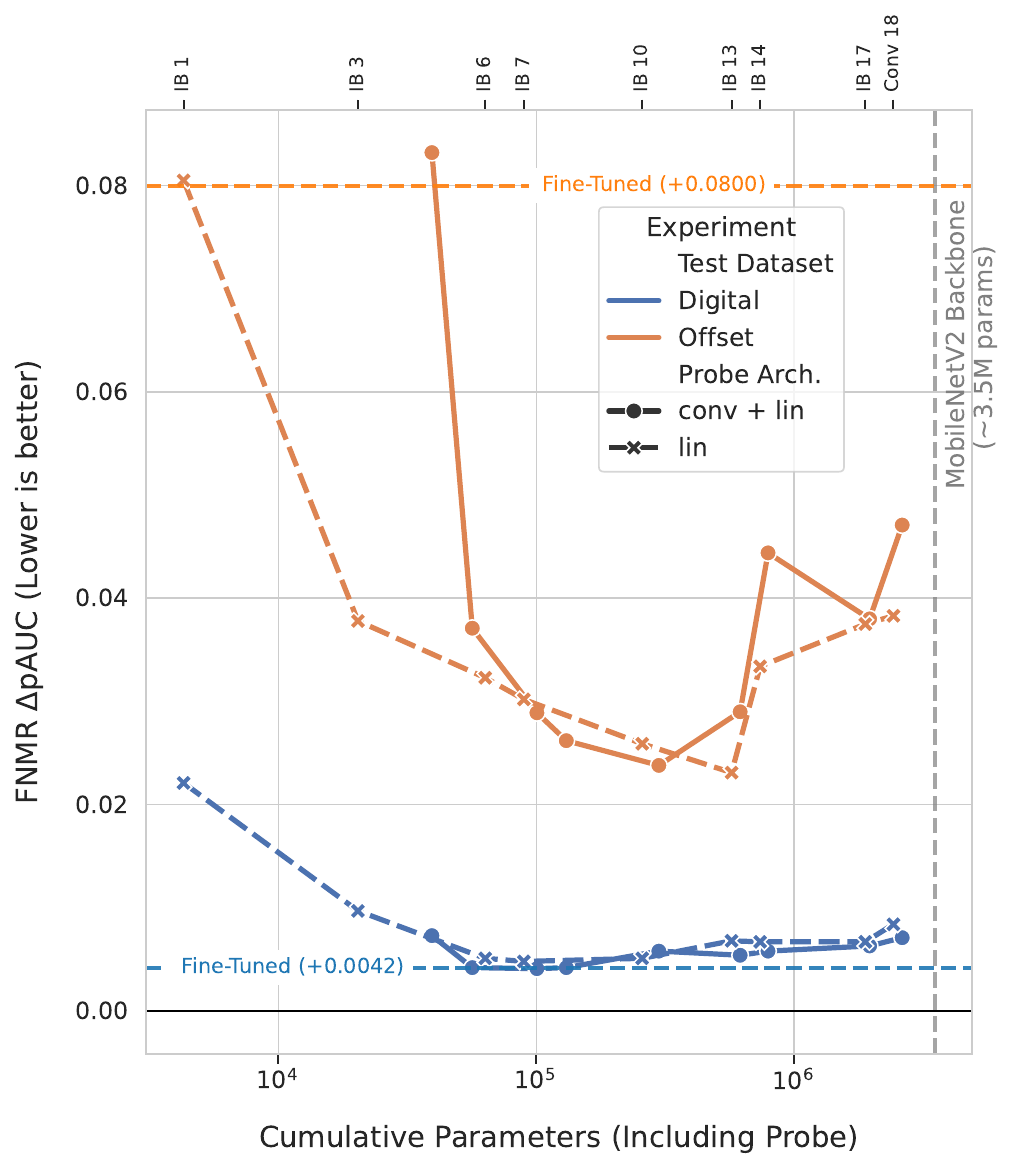}
\vspace{-5mm}
\caption{
    \textbf{Probing Reveals a Trade-off Between In-Domain Specialization and Cross-Domain Generalization.}
    Performance of lightweight probes attached to different layers of a frozen \texttt{MobileNetV2} backbone, with deeper probes resulting in higher parameter counts (x-axis, log scale). The y-axis shows excess error (FNMR $\Delta$pAUC), over the 0-70\% discard rate range. For the \textbf{In-Domain} digital print (blue), performance is optimal in the earlier layers, and the fully \textbf{Fine-Tuned} model (dashed blue line) yields the best performance overall. For the \textbf{Cross-Domain} offset print (orange), however, all probes from IB 3 onwards decisively outperform the fine-tuned model. This demonstrates that full fine-tuning overfits to source-domain artifacts, while probing a general-purpose backbone yields a more robust model for unseen physical domains.
}
\label{fig:probe_results}
\vspace{-4mm}
\end{figure}

\paragraph{Motivation.} A practical system must generalize to varied acquisition conditions and physical manufacturing processes.
\vspace{-.2cm}
\paragraph{Setup.} We compare two approaches: the fully fine-tuned model and a lightweight probe on a frozen backbone. \Cref{fig:probing_methodology} illustrates the core principle of this experimental setup. We take the models trained \emph{only} on the digital press data and evaluate their zero-shot performance on the cross-domain test set.
\vspace{-.2cm}
\paragraph{Results and Discussion.} As expected, all models degrade on the unseen domain. However, the probing analysis reveals a striking and counter-intuitive finding, shown in \Cref{fig:probe_results}.

For the \textbf{in-domain task}, probes on \textbf{early-to-mid layers} perform best. For the \textbf{cross-domain task}, however, a probe on \textbf{middle, more abstract layers} decisively outperforms other probes and even the \textbf{fully fine-tuned model}. This is our key finding: \emph{full fine-tuning overfits to microscopic physical artifacts of the source printing process.} Allowing all layers to adapt makes the model too specific, hurting generalization. In contrast, a probe on a frozen, high-level feature representation leverages general features from ImageNet without corruption. This reveals a critical trade-off between specialization and robustness, an important insight for building generalizable vision systems for the physical world. Full numerical data is in the supplement.


\section{Conclusion}
\label{sec:conclusion}

This paper addressed the ``minimum quality gap'' that hinders widespread adoption of robust mobile verification for physical security graphics. We introduced proactive, predictive quality assessment, where a real-time on-device model acts as an intelligent gatekeeper for a verification system. We proposed a principled, biometrics-inspired framework for evaluating such systems, focusing on the trade-off between verification robustness (FNMR) and usability (ISRR).

Our experiments on a large-scale, diverse dataset validate our approach, showing that a fine-tuned \texttt{MobileNetV2} can effectively predict a definitive quality score. Crucially, our cross-domain analysis revealed that for systems interacting with the physical world, full fine-tuning can overfit to manufacturing artifacts; a more robust model is achieved by probing features from a frozen backbone. This work provides a framework for more reliable physical-digital systems and a key insight into real-world generalization.
\vspace{-.2cm}
\paragraph{Limitations.} Our study was limited to two printing technologies (digital and offset) and, for the cross-domain analysis, a single secure graphic design. Furthermore, our architectural evaluation centered on the widely adopted \texttt{MobileNetV2}. While this demonstrates the feasibility of our approach, we acknowledge that the exploration of other modern efficient architectures could yield further improvements. Future work should incorporate a wider array of production methods (e.g., flexography, laser engraving) and substrates to build a more universal quality model.
\vspace{-.2cm}
\paragraph{Future Work.} A promising future direction lies in leveraging Multimodal Large Language Models (MLLMs) to evolve beyond a single predictive score towards providing granular, actionable user feedback. A practical two-stage roadmap involves first using MLLMs as an automated engine to annotate a large dataset with fine-grained quality defect labels (e.g., \texttt{motion\_blur}, \texttt{glare}) \cite{You2024Descriptive, Wu2024Comprehensive}. Another avenue is the investigation of newer mobile-first architectures \cite{Li2023Rethinking}, which could offer a better trade-off between accuracy and on-device latency. A rich dataset can then be used to distill knowledge into a compact, efficient, multi-task student model suitable for on-device deployment. This process can be enhanced by using few-shot in-context learning, inspired by recent work in related security domains \cite{Komaty2025Exploring}, to generate more accurate labels for the distillation pipeline, creating a path towards truly cooperative verification systems.


\iftoggle{iccvfinal}{
\section*{Acknowledgements}
The authors of the paper wish to acknowledge Justin Picard, Michael Bolay, and Hemang Chawla for the provided experimental infrastructure and insightful discussions. This work was carried out as part of Cas Steigstra’s Master’s thesis in the Artificial Intelligence program at the University of Amsterdam, completed during his internship at Scantrust.}

{
\small
\bibliographystyle{ieeenat_fullname}
\bibliography{main}
}

\clearpage
\appendix
\renewcommand{\thesection}{S\arabic{section}}
\renewcommand{\thefigure}{S\arabic{figure}}
\renewcommand{\thetable}{S\arabic{table}}
\renewcommand{\theequation}{S\arabic{equation}}

\twocolumn[
\begin{center}
    \section*{\LARGE Supplementary Material}
    \vspace{3em}
\end{center}
]

\section{Overview}
This supplementary document provides additional details, figures, and tables to complement our main paper, "Predictive Quality Assessment for Mobile Secure Graphics." The content herein is intended to offer a deeper insight into our methodologies and provide the complete, unabridged results of our empirical evaluation.

\section{Detailed Model Descriptions}
Here, we provide further implementation details for each of the model paradigms evaluated in our work.

\paragraph{Baselines.}
These simple heuristics establish a performance floor. The \texttt{Random} model draws a score from a uniform distribution $Q_{random} = U(0, 1)$. The \texttt{Sharpness} metric is the weighted average of gradient magnitudes above an Otsu-derived threshold: $Q_{\text{sharpness}}(I_S) = (\sum_{g > \tau_{otsu}} g \cdot H(g)) / (\sum_{g > \tau_{otsu}} H(g))$, where $H(g)$ is the gradient histogram. The \texttt{Blur} metric is a specialized measure of effective edge width relative to the QR code's cell size: $Q_{\text{blur}}(I_S) = u / (t \cdot csp)$, where $u$ is the number of gray pixels, $t$ is the number of black-to-white transitions, and $csp$ is the cell size in pixels.

\paragraph{General-Purpose IQA Models.}
The \texttt{BRISQUE\textsuperscript{NSS}} model uses the standard implementation from \cite{Mittal2012No}. It extracts a 36-dimensional feature vector based on the distribution of Mean Subtracted Contrast Normalized (MSCN) coefficients and feeds them into a pre-trained SVR. The \texttt{CLIP-IQA} models use the ViT-L/14 version of CLIP \cite{Radford2021Learning}. The semantic version uses prompts ``Good photo'' and ``Bad photo''. The attribute-based version uses engineered prompts: ``A high-resolution scan of a qr-code with crisp, clear, distinct details'' (positive) and ``A low-resolution scan of a qr-code with blurry, washed-out, indistinct details'' (negative). The score is the softmax probability of similarity to the positive prompt.

\paragraph{Unsupervised (Task-Adapted) Models.}
The \texttt{NIQE (SG)} model adapts the NIQE framework \cite{Mittal2013Making} by building a reference Multivariate Gaussian (MVG) model of BRISQUE features from a corpus of high-quality Secure Graphic (SG) scans (those with $M > 0.95$ in our training set). The quality score is the Mahalanobis distance to this reference model. \texttt{NIQE-LBP (SG)} follows the same principle but replaces the NSS-based features with histograms of Local Binary Patterns (LBP), hypothesizing that texture features are more suitable for non-NSS patterns.

\paragraph{Supervised (Handcrafted) Models.}
The \texttt{BRISQUE (SG|M)} model extracts the 36D BRISQUE feature vector for each image in our training set and trains a Support Vector Regressor (SVR) with an RBF kernel to map these features to the ground-truth score $M$. The \texttt{LBP (SG|M)} model uses a more sophisticated feature set based on locally weighted statistics of uniform LBP codes, adapted from \cite{Wu2015Highly}. An extensive hyperparameter search found that a configuration of $P=8$ neighbors at a radius $R=8$ yielded a 10-dimensional feature vector with the best separability, which was then used to train an SVR.

\paragraph{Supervised (End-to-End) Models.}
The shallow \texttt{CNN-3x32 (SG|M)} model consists of three convolutional blocks (3x3 Conv, ReLU, 2x2 MaxPool) followed by a regression head of two fully-connected layers. The \texttt{MobileNet (SG|M)} model uses the \texttt{MobileNetV2} architecture with randomly initialized weights, trained from scratch. Our main proposal, \texttt{MobileNet\textsuperscript{IN}(SG|M)}, uses the \texttt{MobileNetV2} backbone pre-trained on ImageNet, with the top classification layer replaced by our regression head, and the entire network is fine-tuned on our SG dataset.

\begin{table*}[ht!]
    \centering
    \caption{Full unabridged performance results for all models on both test sets. Performance is measured by $\Delta$pAUC over the 0-70\% discard rate range. Lower values are better. Best performance in each column is in \textbf{bold}, with second-best in \textit{italics}.}
    \label{tab:supp_full_results}
    \resizebox{\textwidth}{!}{%
    \begin{tabular}{@{}lcccc@{}}
    \toprule
    \multirow{2}{*}{\textbf{Method}} & \multicolumn{2}{c}{\textbf{Digital (In-Domain)}} & \multicolumn{2}{c}{\textbf{Offset (Cross-Domain)}} \\
    \cmidrule(lr){2-3} \cmidrule(lr){4-5}
    & FNMR $\Delta$pAUC & ISRR $\Delta$pAUC & FNMR $\Delta$pAUC & ISRR $\Delta$pAUC \\
    \midrule
    \textit{Baselines} & & & & \\
    \texttt{Random} & 0.2826 & 0.2967 & 0.2774 & 0.3000 \\
    \texttt{Sharpness} & 0.0412 & 0.0398 & 0.0871 & 0.0894 \\
    \texttt{Blur} & 0.0441 & 0.0414 & 0.0936 & 0.1003 \\
    \addlinespace
    \textit{General-Purpose IQA} & & & & \\
    \texttt{BRISQUE\textsuperscript{NSS}} & 0.3504 & 0.3862 & 0.2243 & 0.2468 \\
    \texttt{CLIP-IQA (Semantic)} & 0.3810 & 0.4003 & 0.3970 & 0.4318 \\
    \texttt{CLIP-IQA (Attribute)}& 0.2601 & 0.2701 & 0.2407 & 0.2530 \\
    \addlinespace
    \textit{Unsupervised (Task-Adapted)} & & & & \\
    \texttt{NIQE (SG)} & 0.0841 & 0.0828 & 0.1010 & 0.1027 \\
    \texttt{NIQE-LBP (SG)} & 0.0273 & 0.0270 & \textbf{0.0367} & \textbf{0.0356} \\
    \addlinespace
    \textit{Supervised (Handcrafted)} & & & & \\
    \texttt{BRISQUE (SG|M)} & 0.0301 & 0.0295 & 0.0574 & 0.0607 \\
    \texttt{LBP (SG|M)} & 0.0185 & 0.0173 & \textit{0.0390} & \textit{0.0411} \\
    \addlinespace
    \textit{Supervised (End-to-End)} & & & & \\
    \texttt{CNN-3x32 (SG|M)} & 0.0086 & 0.0086 & 0.3441 & 0.3662 \\
    \texttt{MobileNet (SG|M)} & \textit{0.0063} & \textit{0.0064} & 0.1765 & 0.1788 \\
    \texttt{MobileNet\textsuperscript{IN}(SG|M)} & \textbf{0.0042} & \textbf{0.0042} & 0.0800 & 0.0788 \\
    \bottomrule
    \end{tabular}
    }
\end{table*}

\section{Full Evaluation Tables}
The main paper presents a summarized version of the key results. \Cref{tab:supp_full_results} provides the complete, unabridged performance results for all evaluated models on both the in-domain (Digital) and cross-domain (Offset) test sets.

\section{Qualitative Results}
\begin{figure*}[ht!]
    \centering
    \includegraphics[width=0.85\textwidth]{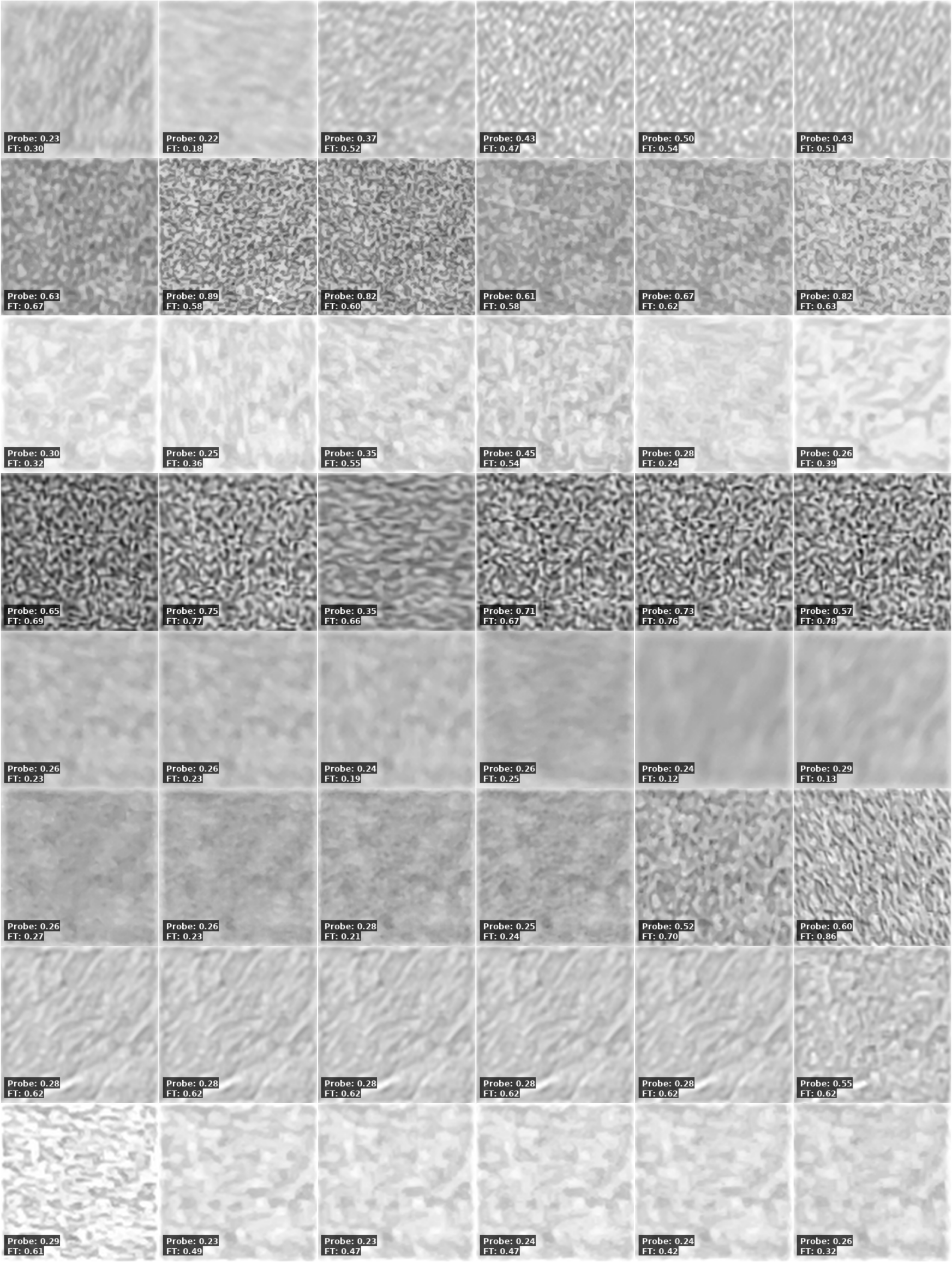} 
    \caption{
    \textbf{Qualitative Comparison of Probe vs. Fine-Tuned (FT) Models on the Cross-Domain (Offset) Test Set.}
    This grid reveals a key failure mode of the FT model. When Offset prints are captured with slight blur or low detail (e.g., rows 3, 4, 7), their texture can incidentally resemble the source (Digital) domain's characteristics. The FT model, having overfitted to these source-domain artifacts, often assigns these defective frames an erroneously high quality score. Conversely, the Probe model proves more robust; it correctly assigns low scores to these degraded frames and, unlike the FT model, reserves its highest scores for the sharpest, highest-fidelity scans (e.g., rows 1 and 3). This highlights the probe's superior generalization in tracking true fidelity.}
    \label{fig:supp_qual_cross_domain}
\end{figure*}

This section provides qualitative examples to visually complement the quantitative results, see \Cref{fig:supp_qual_cross_domain}.

\section{Ablation Study: Network Probing}
This section provides the complete results for the network probing analysis described in Experiment 3 of the main paper.

\subsection{Probe Architectures}
To test different hypotheses about the feature space, we designed two lightweight probe architectures.

\paragraph{Linear Probe (\texttt{lin}).} This simplest probe tests the linear separability of features at a given layer. Its architecture is a Global Average Pooling (GAP) layer followed by a single fully-connected (Dense) layer that maps the feature vector to the final quality score.

\paragraph{\texttt{conv + lin} Probe.} This probe has slightly more capacity. It consists of a $1 \times 1$ Convolutional layer, which acts as a channel-wise feature recombiner, followed by the same GAP and Dense layer structure as the linear probe. This allows the model to learn an optimal linear combination of feature channels before the final regression.

\subsection{Choice of Probe Layer}
\Cref{tab:supp_probe_full} provides the complete numerical results for both probe architectures and the fine-tuned model, attached to all investigated intermediate blocks of the frozen \texttt{MobileNetV2} backbone. This data provides a granular view of where predictive information resides within the network and highlights the trade-off between specialization and generalization, comparing in-domain and cross-domain performance against the fully fine-tuned model.

\begin{table*}[ht!]
    \centering
    \caption{Complete Unabridged Probe Performance vs. Fine-Tuned Model. This table compares probe performance against the fully fine-tuned model. Lower $\Delta$pAUC values are better. The best result in each column is shown in \textbf{bold}.}
    \label{tab:supp_probe_full}
    \resizebox{\textwidth}{!}{%
    \begin{tabular}{@{}ll S[table-format=1.4] S[table-format=1.4] S[table-format=1.4] S[table-format=1.4]@{}}
    \toprule
    \multirow{2}{*}{\textbf{Probe Arch.}} & \multirow{2}{*}{\textbf{Probed Layer}} & \multicolumn{2}{c}{\textbf{Digital (In-Domain)}} & \multicolumn{2}{c}{\textbf{Offset (Cross-Domain)}} \\
    \cmidrule(lr){3-4} \cmidrule(lr){5-6}
    & & {FNMR $\Delta$pAUC} & {ISRR $\Delta$pAUC} & {FNMR $\Delta$pAUC} & {ISRR $\Delta$pAUC} \\
    \midrule
    \multicolumn{6}{c}{\textit{\texttt{conv + lin} Probes}} \\
    \midrule
    \texttt{conv + lin} & IB 1 & 0.0073 & 0.0077 & 0.0832 & 0.0944 \\
    \texttt{conv + lin} & IB 3 & 0.0042 & 0.0046 & 0.0371 & 0.0403 \\
    \texttt{conv + lin} & IB 6 & \textbf{0.0041} & 0.0044 & 0.0289 & 0.0323 \\
    \texttt{conv + lin} & IB 7 & 0.0042 & 0.0046 & 0.0262 & 0.0297 \\
    \texttt{conv + lin} & IB 10 & 0.0058 & 0.0060 & 0.0238 & 0.0269 \\
    \texttt{conv + lin} & IB 13 & 0.0054 & 0.0057 & 0.0290 & 0.0328 \\
    \texttt{conv + lin} & IB 14 & 0.0058 & 0.0065 & 0.0444 & 0.0529 \\
    \texttt{conv + lin} & IB 17 & 0.0063 & 0.0066 & 0.0380 & 0.0431 \\
    \texttt{conv + lin} & Conv 18 & 0.0071 & 0.0074 & 0.0471 & 0.0559 \\
    \midrule
    \multicolumn{6}{c}{\textit{\texttt{lin} Probes}} \\
    \midrule
    \texttt{lin} & IB 1 & 0.0221 & 0.0218 & 0.0805 & 0.0851 \\
    \texttt{lin} & IB 3 & 0.0097 & 0.0100 & 0.0378 & 0.0407 \\
    \texttt{lin} & IB 6 & 0.0051 & 0.0055 & 0.0323 & 0.0364 \\
    \texttt{lin} & IB 7 & 0.0048 & 0.0052 & 0.0302 & 0.0325 \\
    \texttt{lin} & IB 10 & 0.0051 & 0.0054 & 0.0259 & 0.0284 \\
    \texttt{lin} & IB 13 & 0.0068 & 0.0071 & \textbf{0.0231} & \textbf{0.0255} \\
    \texttt{lin} & IB 14 & 0.0067 & 0.0070 & 0.0334 & 0.0372 \\
    \texttt{lin} & IB 17 & 0.0067 & 0.0070 & 0.0375 & 0.0435 \\
    \texttt{lin} & Conv 18 & 0.0084 & 0.0086 & 0.0383 & 0.0427 \\
    \midrule
    \multicolumn{6}{c}{\textit{Fully Fine-Tuned Model (Reference)}} \\
    \midrule
    \multicolumn{2}{l}{\texttt{MobileNet\textsuperscript{IN}(SG|M)}} & 0.0042 & \textbf{0.0042} & 0.0800 & 0.0788 \\
    \bottomrule
    \end{tabular}
    }
\end{table*}


\end{document}